\documentclass{article} 
\usepackage{nips13submit_e,times}
\usepackage{fancyhdr} 
\usepackage{url}
\usepackage{graphicx}
\usepackage{epsfig}
\usepackage{amssymb}
\usepackage{epstopdf}
\usepackage{multirow}
\usepackage{booktabs}
\usepackage{threeparttable}
\usepackage{tabularx}
\usepackage[hidelinks]{hyperref}
\hypersetup{
    colorlinks,
    citecolor=green,
    filecolor=black,
    linkcolor=red,
    urlcolor=blue
}

\title{Leveraging Mid-Level Deep Representations For Predicting Face Attributes in the Wild}

\author{
Yang Zhong \qquad Josephine Sullivan \qquad Haibo Li \\
Computer Science and Communication\\
KTH Royal Institute of Technology\\
100 44 Stockholm, Sweden\\
\texttt{\{yzhong, sullivan, haiboli\}@kth.se} \\
}

\nipsfinalcopy 

\begin{document}

\clearpage
\maketitle
\thispagestyle{empty}

\begin{abstract}
Predicting facial attributes from faces in the wild is very challenging due to pose and lighting variations in the real world. 
The key to this problem is to build proper feature representations to cope with these unfavourable conditions. 
Given the success of Convolutional Neural Network (CNN) in image classification, the high-level CNN feature, as an intuitive and reasonable choice, has been widely utilized for this problem.
In this paper, however, we consider the mid-level CNN features as an alternative to the high-level ones for attribute prediction.
This is based on the observation that face attributes are different: some of them are locally oriented while others are globally defined.
Our investigations reveal that the mid-level deep representations outperform the prediction accuracy achieved by the (fine-tuned) high-level abstractions. 
We empirically demonstrate that the mid-level representations achieve state-of-the-art prediction performance on CelebA and LFWA datasets.
Our investigations also show that by utilizing the mid-level representations one can employ a single deep network to achieve both face recognition and attribute prediction.
\end{abstract}

\section{Introduction}
\label{sec:intro}

Telling attributes from face images in the wild is known to be very useful in large scale face search, image understanding and face recognition.  
However, the problem indeed is very challenging since the faces captured in the real world are often affected by many unfavourable influences, such as illumination, pose and expression. 
To build usable attribute prediction, it is important to preserve the essential face traits in the representations and at the same time, make them insensitive to interference. 

The representations describing faces in prior literature generally form two groups: the hand-crafted representations and learned representations. 
Exemplified as in \cite{kumar2009attribute, bourdev2011describing}, local low-level features were constructed from detected face regions.
The use of local features were mostly based on the consideration that they are less likely to be influenced than the holistic features by pose changes and facial expressions. 
In recent work \cite{li2015two}, multi-scale Gabor features were used as a holistic face representation, which is then converted by a learned hashing process for efficient face retrieval and attribute prediction.
 
Driven by the great improvements brought by the CNN in image classification \cite{krizhevsky2012imagenet, szegedy2014going, simonyan2014very} and face recognition \cite{sun2013deep, parkhi2015deep, taigman2013deepface, schroff2015facenet}, features extracted from deep architectures became a natural and reasonable choice to represent faces for attribute prediction.
In \cite{6909608}, local semantic image patches were first detected and fed into deep networks to construct concatenated, pose normalized representation. 
In \cite{liu2015faceattributes}, through intensive training, two concatenating CNNs were built to locate and predict attributes from arbitrary size of faces in the wild.
The first CNN stage was pre-trained by image categories and fine-tuned by face attribute tags to locate face from complex background; 
the second stage was pre-trained by identity labels and fine-tuned by face attribute data 
to achieve an effective fusion of the discrimination of inter-person representations and the variability of non-identity related factors in the face representations.
As a common character, high-level hidden (Fully Connected, FC) layers were especially trained for representing attributes in these work. 

The best representation for retrieval tasks has been shown to come from the FC layer \cite{azizpour2015generic,  razavian2014cnn}.
Naturally, features from FC layers are commonly used for attribute prediction.  
But considering that different levels of ConvNet encode different levels of abstraction, one can expect that such representations may not be optimal to describe the physical facial characteristics, especially the local attributes, such as ``open mouth'' and ``wearing glasses''.
It is therefore rational to consider CNN features from earlier layers, which better preserve both discriminating power and spatial information, for predicting face attributes. 

As indicated in \cite{FaceAttriOffShelf}, the earlier-layer CNN features are likely to better describe the facial appearance than the high-level features.
In this work, our focus is on identifying the best face representations for recognizing face attributes.  
We employ publicly available data, architecture and a deep learning framework to train a face classification CNN.
Hierarchical representations are then constructed and evaluated in a face attribute prediction context.
Through intensive investigations, we empirically show the effectiveness of the hierarchical deep representations and  demonstrate the advantages of the mid-level representations for tackling the face attribute prediction problem.  
The major contributions of this work are:
\begin{itemize} 
  \item Our investigations reveal the fact of the diverse utilities of deep hierarchical representations for face attribute prediction; intermediate representations are shown very effective for predicting the human describable face attributes (Section \ref{sec:res}).
  \item  By jointly leveraging the mid-level hierarchical representations, we further improve the state-of-the-art on two large scale datasets (Section \ref{sec:bench}).
  \item One can construct a single deep network for both face recognition and attribute prediction.
\end{itemize}

\section{Investigation}
\label{sec:ivst}
Our motivation in this work is to study the effectiveness of hierarchical CNN representations for attribute prediction of faces captured in the wild. 
Our procedure is to first extract hierarchical representations from aligned faces using our trained CNN and then we construct and evaluate linear attribute classifiers to identify the most effective representation of each attribute. 
In the following sections, we first describe the experimental details and then present our investigations and results on the CelebA dataset and the annotated LFW (LFWA) dataset.
  
\subsection{Implementation details}
\label{details}

\textbf{CNN Training}: The face representations studied in this work were extracted from a face classification CNN.
The architecture of the CNN with configuration details is shown in Table \ref{tab:CNN}.
We adopted the structure of the convolutional filters of ``FaceNet NN.1 \cite{schroff2015facenet}'' , with slight modifications, into our CNN and concatenated two $512d$ FC layers and trained it in a N-way classification manner.
The network was initialized from random and started with a learning rate of $0.015$, which was then decreased two times when the classification accuracy stopped increasing on the validation set.
PReLU \cite{he2015delving} rectification was used after all the convolutional and FC layers. Dropout layers were inserted between FC layers with dropout rate of $0.5$ to prevent overfitting. 
We used around $10000$ identities with $350000$ face images from the publicly accessible dataset WebFace \cite{yi2014learning} for training.
The training instances were augmented with random flipping, jittering, and slight rotation.
The trained CNN had a verification accuracy of $97.5\%$ on LFW dataset \cite{Huang2007}, which is comparable to that of the DeepID \cite{sun2013deep} structure used by \cite{liu2015faceattributes}.  

\begin{table}[]
\centering
\small
\caption{CNN architecture used in our experiments. The perception size is 112 $\times$ 112 (except experiments in Section \ref{sec:discuss}). The dimensions of the investigated deep representations are shown in the last column.}
\label{tab:CNN}
\begin{tabular}{cccc}
\hline
\toprule
Layer   & Kernel     & Output    & \begin{tabular}[c]{@{}c@{}}Representation\\ Dimension\end{tabular} \\ \hline
Conv 1   & conv3-64   &           &                                                                    \\ \hline
Pool 1  & max 2,2    & 56*56*64  &                                                         \\ \hline
RNorm 1 & 5          &           &                                                                    \\ \hline
Conv 2a & conv1-64   &           &                                                                    \\ \hline
Conv 2   & conv3-192  & 28*28*192 & 3* 3* 192                                                          \\ \hline
RNorm 2  & 5          &           &                                                                    \\ \hline
Pool 2   & max 2,2    &           &                                                                    \\ \hline
Conv 3a  & conv1-192  &           &                                                                    \\ \hline
Conv 3  & conv3-384  & 14*14*384 & 3* 3* 384                                                          \\ \hline
Pool 3   & max 2,2    &           &                                                                    \\ \hline
Conv 4a  & conv1-384  &           &                                                                    \\ \hline
Conv 4   & conv3-256  & 14*14*384 & 3* 3* 384                                                          \\ \hline
Conv 5a & conv1-256  &           &                                                                    \\ \hline
Conv 5  & conv3-256  & 14*14*384 & 3* 3* 384                                                          \\ \hline
Conv 6a & conv1-256  &           &                                                                    \\ \hline
Conv 6  & conv3-256 & 7*7*256   & 3* 3* 256                                                          \\ \hline
Pool 6   &            &           &                                                                    \\ \hline
FC 1    & 512        & 1*1*512   &                                                                    \\ \hline
FC 2    & 512        & 1*1*512   &                                                                    \\ 
\bottomrule
\hline
\end{tabular}
\end{table}

\textbf{Feature Representations}: 
In our experiments faces were aligned based on the detected feature points \cite{dlib09} (or provided feature points). 
The aligned image, approximately covering from the top of head to neck, had a size of $120 \times 120$.
The center patch, with a size of $112 \times 112$, and its horizontal flipped patch, were fed in to the CNN.
We extracted the average representations from $Conv2$, $Conv3$, $Conv4$, $Conv5$, $Conv6$, $FC1$ and $FC2$ layers, and additionally applied multi-scale spatial pooling to improve the invariance of CNN activations without degrading their discriminating power.
The first average pooling halved the dimension and the second overlapping max.\ pooling selected the most locally activated neuron\footnote{For example, for $Conv2$ layer, the first stage is  average pooling (window = 4 $\times$ 4, stride = 4) and the second one is max. pooling (window = 3 $\times$ 3, stride = 2).}. 
That is, all the resulting representations from the convolutional layers had the same feature map size of 3 $\times$ 3 as shown in the last column of Table \ref{tab:CNN}.
The resulting features are correspondingly denoted by  $C2$ , ... , $C6$, $F1$  and $F2$ in the following.

\textbf{Evaluation Dataset}: 
In this work, we used two datasets\footnote{\url{http://mmlab.ie.cuhk.edu.hk/projects/CelebA.html}. Released in Oct 2015.}, CelebA and LFWA, to conduct benchmark evaluations of attribute prediction accuracy.
The CelebA contains around $200,000$ images of $10,000$ identities and LFWA has $13,233$ images of $5,749$ identities. 
Images on both datasets are labeled with 40 binary codes to represent the presence of facial attributes. 
These attributes ranges from demographic information to physical characteristics such as nose size, eyebrow shape and expressions.
  
\subsection{Exploration}
\label{sec:res}

\begin{figure*}[t!]
\begin{center}
   \includegraphics[width=1\linewidth]{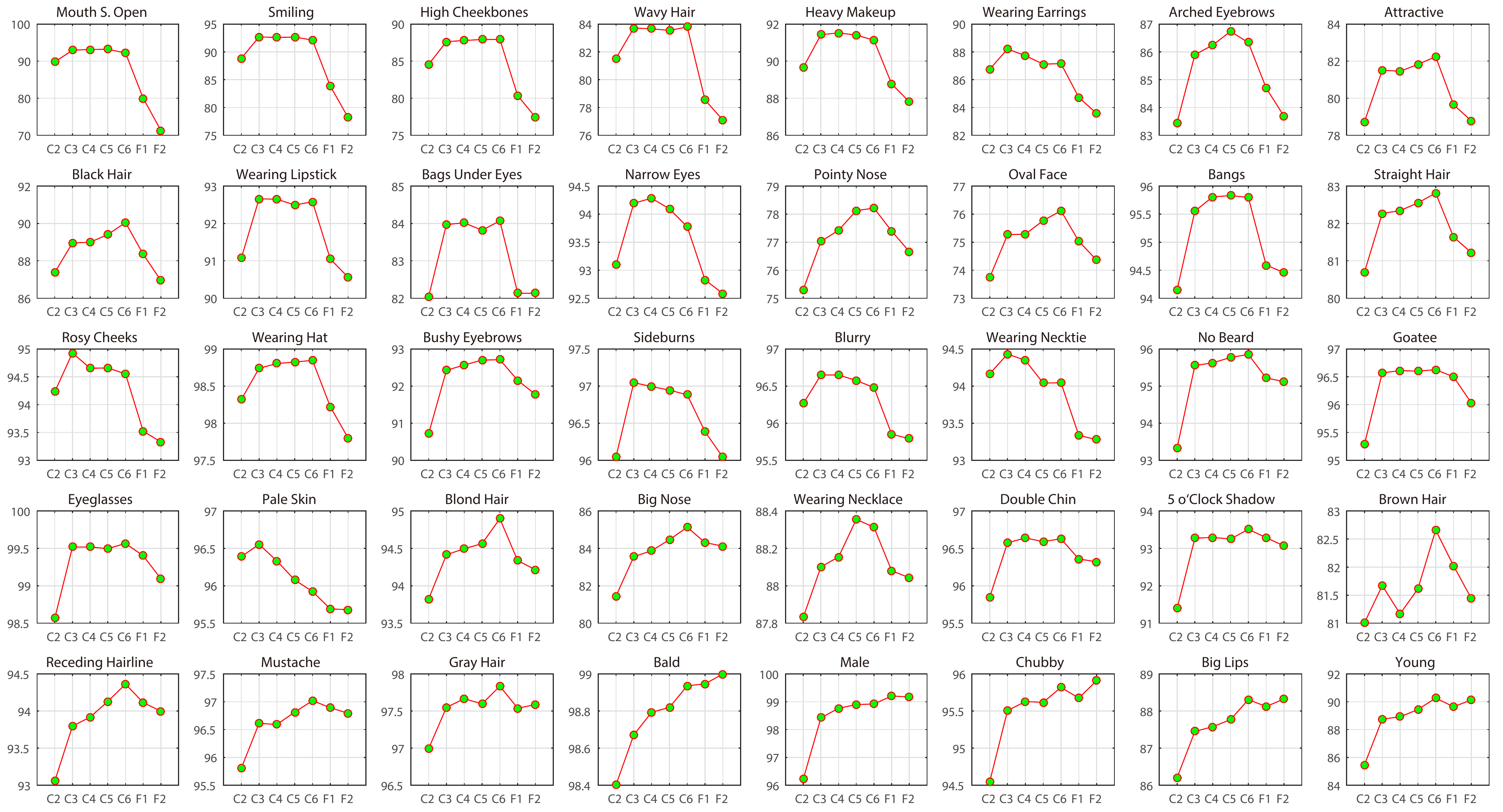}
\end{center}
   \caption{Prediction effectiveness discrepancy of hierarchical representations over 40 attributes on the training set of CelebA. In each grid, y-scale stands for the prediction accuracy in \%. The mean prediction accuracy values of each representation (from $C2$ - $F2$) are $89.8 \%$ , $91.3 \%$, $91,3 \%$, $91.4 \%$, $91.5 \%$, $90.0 \%$ and $89.2 \%$.} 
\label{fig:perfDiff}
\end{figure*}

The attribute prediction power of the hierarchical representations was first investigated on the training set of the CelebA dataset.
We used the training set defined in \cite{liu2015faceattributes} to construct linear SVM attribute classifiers for each representation. 
The prediction accuracy on all the 40 attributes for each type of representation is demonstrated in Figure \ref{fig:perfDiff}.  
It can be observed that features from the intermediate layers ($Conv3$ to $Conv6$) demonstrate an obvious advantage over the final FC layer on average, especially for attributes describing motions of the mouth area where the gap is almost $20\%$.  
Feature $C6$ had the highest prediction accuracy on average, which was more than $2\%$ higher than the last FC layer. 
One can also observe that the mid-level convolutional representations also featured slightly different prediction power on different attributes, e.g.\ $C2$ outperformed on ``Rosy Cheeks'' but not on hair related attributes.
Through this exploration, we identified the best representation for each attribute.

\subsection{Benchmark evaluations}
\label{sec:bench}
\begin{figure*}[!]
\begin{center}
   \includegraphics[width=0.8\linewidth]{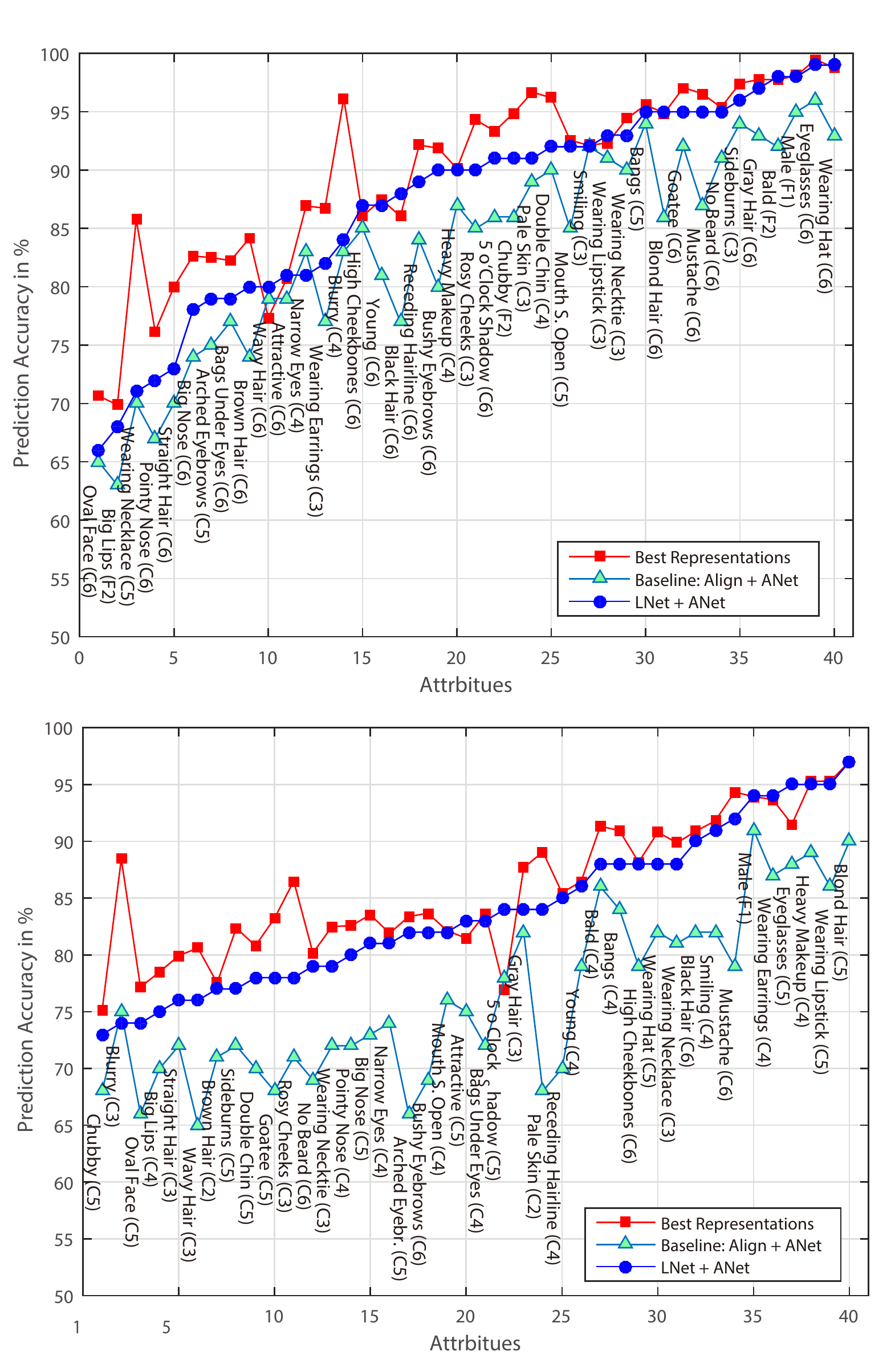}
\end{center}
   \caption{Comparative prediction accuracy on CelebA (upper) and LFWA (lower) with best performing representation given in parenthesis after each attribute (refer our numeric accuracy values to Table \ref{tab:combine}). }
\label{fig:bothDB}
\end{figure*}

We then compared the identified best representations of each attribute with a baseline approach and the current state-of-the-art \cite{liu2015faceattributes}, on CelebA and LFWA datasets respectively.
The results are compared in Figure \ref{fig:bothDB} and the detailed recognition rates of our approach are given in Table \ref{tab:combine}.
The baseline approach (denoted by [17]+ ANetin in \cite{liu2015faceattributes}) used CNN trained by identities to extract features from aligned faces; the pipeline of constructing deep representations is the same as ours.  
The current best, denoted by ``LNet+ANet'' in \cite{liu2015faceattributes}, employed concatenated CNNs to locate faces and extract features in order to construct attribute classifiers. 

One can observe that: by jointly leveraging the best representations, superior performance has been achieved over the equivalent baseline approach and even the current state-of-the-art. 
For some attributes with relatively lower (baseline) accuracy, e.g.\ ``Blurry'' ``Oval Face'', ``Wearing Necktie'' and ``Rosy Cheeks'', the advantage of our approach is especially significant. 
Among the $40$ evaluated attributes, the mid-level convolutional features dominated the best representations (only 4 best representations came from FC layers on CelebA and 1 on LFWA).

In addition, we also comprehensively compared the overall performance in Table \ref{tab:compare} and layer wise prediction accuracy in Table \ref{tab:furtherinfo}. 
It can be found that: 
\begin{enumerate}
\item utilizing the best representations demonstrates \textbf{outperforming effectiveness} over the equivalent baseline method and state-of-the-art approach on both large scale datasets; 

\item the prediction power increases along the convolutional layers of the CNN;

\item even the very early layer $C2$ astonishingly achieved entirely comparable performance as the state-of-the-art.

\end{enumerate}
Indeed, even if we utilize convolutional features on the attributes which had FC as the best representations (e.g.\ ``Bald''), the effect on prediction performance is not detectable (which also reflects in Figure \ref{fig:perfDiff}).

\begin{table}[]
\centering
\small
\caption{Our attribute recognition accuracy on the CelebA and LFWA in \%}
\label{tab:combine}
\begin{tabular}{lrr|lrr}
\hline
                 & CelebA & LFWA  &                   & CelebA & LFWA  \\ \hline
5 o'Clock Shadow & 93.34  & 76.90 & Male              & 98.09  & 93.93 \\ \hline
Arched Eyebrows  & 82.50  & 83.37 & Mouth S. Open     & 92.56  & 82.13 \\ \hline
Attractive       & 80.77  & 81.47 & Mustache          & 96.56  & 94.33 \\ \hline
Bags Under Eyes  & 82.24  & 83.63 & Narrow Eyes       & 86.92  & 81.90 \\ \hline
Bald             & 97.75  & 91.33 & No Beard          & 95.38  & 80.17 \\ \hline
Bangs            & 95.58  & 90.93 & Oval Face         & 70.63  & 77.17 \\ \hline
Big Lips         & 69.90  & 78.50 & Pale Skin         & 96.69  & 89.07 \\ \hline
Big Nose         & 82.64  & 83.53 & Pointy Nose       & 76.17  & 82.57 \\ \hline
Black Hair       & 86.04  & 90.90 & Receding Hairline & 92.14  & 85.47 \\ \hline
Blond Hair       & 94.89  & 97.00 & Rosy Cheeks       & 94.29  & 86.50 \\ \hline
Blurry           & 96.15  & 88.43 & Sideburns         & 97.44  & 82.37 \\ \hline
Brown Hair       & 84.15  & 77.53 & Smiling           & 92.11  & 91.80 \\ \hline
Bushy Eyebrows   & 91.89  & 83.67 & Straight Hair     & 80.00  & 79.93 \\ \hline
Chubby           & 94.87  & 75.13 & Wavy Hair         & 77.35  & 80.60 \\ \hline
Double Chin      & 96.19  & 80.80 & Wearing Earrings  & 86.74  & 93.67 \\ \hline
Eyeglasses       & 99.48  & 91.50 & Wearing Hat       & 98.78  & 90.80 \\ \hline
Goatee           & 97.07  & 83.23 & Wearing Lipstick  & 92.35  & 95.30 \\ \hline
Gray Hair        & 97.77  & 87.67 & Wearing Necklace  & 85.78  & 89.87 \\ \hline
Heavy Makeup     & 90.14  & 95.27 & Wearing Necktie   & 94.42  & 82.50 \\ \hline
High Cheekbones  & 86.06  & 88.17 & Young             & 87.48  & 86.47 \\ \hline
\end{tabular}
\end{table}


Moreover, it is worth nothing that using features at FC layers is computationally expensive due to the bottle neck between convolutional and FC layer.
Thus, leveraging the mid-level representations features two inherent advantages: 
\begin{enumerate}
\item very \textbf{efficient} inference with competitive performance; 

\item breaking the bounds of the interconnections between convolutional and FC layers so that the CNN accepts arbitrary receptive size (as demonstrated in Section \ref{sec:discuss}).
\end{enumerate}

Recall that the training effort of a face classification CNN is almost minimal compared to state-of-the-art two-stage CNN approach and other potential end-to-end learning for attribute prediction.
However, the deep hierarchical representations from the trained face classification CNN still allow superior performance. 
This indicates that these mid-level CNN representations have contained rich spatial information which can be utilized more in building effective (local) appearances description, attribute prediction and retrieval applications.

\begin{table}[]
\centering
\small
\caption{Overall comparisons of the baseline approach, current state-of-the-art (LNet+ANet) and our approach.}
\label{tab:compare}
\begin{tabular}{cccc}
\hline
\toprule
       & Baseline & LNet+ANet & Ours  \\ \hline 
CelebA & 83\%     & 87\%  & \textbf{89.8\%}   \\ \hline 
LFWA   & 76\%     & 84\%  & \textbf{85.9\%}  \\  
\bottomrule
\hline
\end{tabular}
\end{table}

\begin{table}[]
\centering
\small
\caption{Layer wise average prediction accuracy over 40 attributes on the test set of CelebA and LFWA of our approach.}
\label{tab:furtherinfo}
\begin{tabular}{cccccccc}
\hline
\toprule
       & C2   & C3   & C4   & C5   & C6   & F1   & F2   \\ \hline
CelebA & 89\% & 90\% & 90\% & 90\% & 90\% & 87\% & 88\% \\ \hline
LFWA   & 83\% & 86\% & 86\% & 86\% & 85\% & 82\% & 81\% \\ 
\bottomrule
\hline
\end{tabular}
\end{table}

\subsection{Discussion} 
\label{sec:discuss}
 
It is easy to expect that a number of factors, e.g.\ image resolution, could affect the face attribute prediction performance.
A small receptive size would affect the prediction relating to small scale face traits. 
We found that by doubling the input size of our CNN the prediction accuracy for ``Bag under eyes'' increased from $82.2\%$ to $83.5\%$  with $C3$ as the best representation.

Similarly, attributes that are related to the face components with deterministic distribution can have condensed representations.
For instance, ``arched eyebrow'' always appears in the upper part of face images and ``wearing necktie'' appears in the lower part. 
We studied the condensed $C6$ feature ($3\times2\times256$) as the representation for locally related attributes and found that the average prediction accuracy remained almost the same (89.3\%). 
This means that we can further reduce the memory footprint of the representations and the classifiers for the attributes defined by deterministic facial components.

\section{Conclusions}

In this paper, we have proposed to leverage the mid-level representations from deep convolutional neural network to tackle the attribute prediction problem for faces in the wild.
We used an off-the-shelf architecture and a publicly available dataset to train a plain classification network, and conduct investigations on the utility of the deep representations from various levels of the network. 
Although the trained network is not optimized either towards face attribute prediction or recognition, 
it still allows accurate attribute prediction surpassing the state-of-the-art with a noticeable margin. 

Our investigations indicate that CNNs trained for face classification have implicitly learned many semantic concepts and human describable attributes have been embedded in the mid-level deep representations, which can be separated by simple classifiers.
They also reveal the potential utility of the intermediate deep representations for other face related tasks.
These findings promise to increase the face attribute prediction performance and to achieve multiple intelligent functions, such as face recognition, retrieval and attribute prediction, in one single deep architecture.


\bibliographystyle{plain}
\bibliography{ref}

\begin{thebibliography}{10}

\bibitem{azizpour2015generic}
Hossein Azizpour, Ali Razavian, Josephine Sullivan, Atsuto Maki, and Stefan
  Carlsson.
\newblock From generic to specific deep representations for visual recognition.
\newblock In {\em Proceedings of the IEEE Conference on Computer Vision and
  Pattern Recognition Workshops}, pages 36--45, 2015.

\bibitem{bourdev2011describing}
Lubomir Bourdev, Subhransu Maji, and Jitendra Malik.
\newblock Describing people: A poselet-based approach to attribute
  classification.
\newblock In {\em Computer Vision (ICCV), 2011 IEEE International Conference
  on}, pages 1543--1550. IEEE, 2011.

\bibitem{he2015delving}
Kaiming He, Xiangyu Zhang, Shaoqing Ren, and Jian Sun.
\newblock Delving deep into rectifiers: Surpassing human-level performance on
  imagenet classification.
\newblock {\em arXiv preprint arXiv:1502.01852}, 2015.

\bibitem{Huang2007}
Gary~B. Huang, Marwan Mattar, Tamara Berg, and Erik Learned-miller.
\newblock E.: Labeled faces in the wild: A database for studying face
  recognition in unconstrained environments.
\newblock Technical report, 2007.

\bibitem{dlib09}
Davis~E. King.
\newblock Dlib-ml: A machine learning toolkit.
\newblock {\em Journal of Machine Learning Research}, 10:1755--1758, 2009.

\bibitem{krizhevsky2012imagenet}
Alex Krizhevsky, Ilya Sutskever, and Geoffrey~E Hinton.
\newblock Imagenet classification with deep convolutional neural networks.
\newblock In {\em Advances in neural information processing systems}, pages
  1097--1105, 2012.

\bibitem{kumar2009attribute}
Neeraj Kumar, Alexander~C Berg, Peter~N Belhumeur, and Shree~K Nayar.
\newblock Attribute and simile classifiers for face verification.
\newblock In {\em Computer Vision, 2009 IEEE 12th International Conference on},
  pages 365--372. IEEE, 2009.

\bibitem{li2015two}
Yan Li, Ruiping Wang, Haomiao Liu, Huajie Jiang, Shiguang Shan, and Xilin Chen.
\newblock Two birds, one stone: Jointly learning binary code for large-scale
  face image retrieval and attributes prediction.
\newblock In {\em Proceedings of the IEEE International Conference on Computer
  Vision}, pages 3819--3827, 2015.

\bibitem{parkhi2015deep}
Omkar~M Parkhi, Andrea Vedaldi, and Andrew Zisserman.
\newblock Deep face recognition.
\newblock {\em Proceedings of the British Machine Vision}, 1(3):6, 2015.

\bibitem{razavian2014cnn}
Ali~S Razavian, Hossein Azizpour, Josephine Sullivan, and Stefan Carlsson.
\newblock Cnn features off-the-shelf: an astounding baseline for recognition.
\newblock In {\em Computer Vision and Pattern Recognition Workshops (CVPRW),
  2014 IEEE Conference on}, pages 512--519. IEEE, 2014.

\bibitem{schroff2015facenet}
Florian Schroff, Dmitry Kalenichenko, and James Philbin.
\newblock Facenet: A unified embedding for face recognition and clustering.
\newblock In {\em Proceedings of the IEEE Conference on Computer Vision and
  Pattern Recognition}, pages 815--823, 2015.

\bibitem{simonyan2014very}
Karen Simonyan and Andrew Zisserman.
\newblock Very deep convolutional networks for large-scale image recognition.
\newblock {\em arXiv preprint arXiv:1409.1556}, 2014.

\bibitem{sun2013deep}
Yi~Sun, Xiaogang Wang, and Xiaoou Tang.
\newblock Deep learning face representation from predicting 10,000 classes.
\newblock In {\em Proceedings of the IEEE Conference on Computer Vision and
  Pattern Recognition}, pages 1891--1898, 2013.

\bibitem{szegedy2014going}
Christian Szegedy, Wei Liu, Yangqing Jia, Pierre Sermanet, Scott Reed, Dragomir
  Anguelov, Dumitru Erhan, Vincent Vanhoucke, and Andrew Rabinovich.
\newblock Going deeper with convolutions.
\newblock {\em arXiv preprint arXiv:1409.4842}, 2014.

\bibitem{taigman2013deepface}
Yaniv Taigman, Ming Yang, Marc'Aurelio Ranzato, and Lior Wolf.
\newblock Deepface: Closing the gap to human-level performance in face
  verification.
\newblock In {\em Proceedings of the IEEE Conference on Computer Vision and
  Pattern Recognition}, pages 1701--1708, 2013.

\bibitem{yi2014learning}
Dong Yi, Zhen Lei, Shengcai Liao, and Stan~Z Li.
\newblock Learning face representation from scratch.
\newblock {\em arXiv preprint arXiv:1411.7923}, 2014.

\bibitem{6909608}
Ning Zhang, M.~Paluri, M.~Ranzato, T.~Darrell, and L.~Bourdev.
\newblock Panda: Pose aligned networks for deep attribute modeling.
\newblock In {\em Computer Vision and Pattern Recognition (CVPR), 2014 IEEE
  Conference on}, pages 1637--1644, June 2014.

\bibitem{liu2015faceattributes}
Ziwei~Liu, Ping~Luo, Xiaogang~Wang and Xiaoou Tang.
\newblock Deep learning face attributes in the wild.
\newblock In {\em Proceedings of International Conference on Computer Vision
  (ICCV)}, 2015.

\bibitem{FaceAttriOffShelf}
Yang Zhong, Josephine Sullivan, and Haibo Li,
\newblock ``Face Attribute Prediction Using Off-the-Shelf CNN Features,''
\newblock In {\em Proceedings of International Conference on Biometrics
  (ICB)}, 2016.

\end{thebibliography}

\end{document}